# The Arabic Ontology – An Arabic Wordnet with Ontologically Clean Content

*Mustafa Jarrar*
*Birzeit University, Palestine*
*mjarrar@birzeit.edu*

**Abstract**. We present a formal Arabic wordnet built on the basis of a carefully designed ontology hereby referred to as the *Arabic Ontology*. The ontology provides a formal representation of the concepts that the Arabic terms convey, and its content was built with ontological analysis in mind, and benchmarked to scientific advances and rigorous knowledge sources as much as this is possible, rather than to only speakers' beliefs as lexicons typically are. A comprehensive evaluation was conducted thereby demonstrating that the current version of the top-levels of the ontology can top the majority of the Arabic meanings. The ontology consists currently of about 1,300 well-investigated concepts in addition to 11,000 concepts that are partially validated. The ontology is accessible and searchable through a lexicographic search engine (http://ontology.birzeit.edu) that also includes about 150 Arabic-multilingual lexicons, and which are being mapped and enriched using the ontology. The ontology is fully mapped with Princeton WordNet, Wikidata, and other resources.
Accepted by:

## 1. Introduction

The importance of linguistic ontologies and wordnets is increasing in many application areas, such as multilingual big data (Oana et al., 2012; Ceravolo, 2018), information retrieval (Abderrahim et al., 2013), question-answering and NLP-based applications (Shinde et al., 2012), data integration (Castanier et al., 2012; Jarrar et al., 2011), multilingual web (McCrae et al., 2011; Jarrar, 2006), among others. After developing the Princeton English WordNet[1] (Miller et al., 1990), there have been great efforts towards building wordnets for many languages. This trend is a natural evolution of thesauri engineering. A thesaurus, typically, is a list of words classified as near-synonyms or it can be seen as pairs of terms connected through "RelatedTo" and/or "Broader/Narrow" relations. Unfortunately, such semantically-poor and imprecise relationships between words are not sufficient for most IT-based applications (Kless et al., 2012). WordNet was proven to be more useful as it provides more precise and semantically richer results (Castanier et al., 2012; Scott & Matwin, 1998; Huang et al., 2009). A wordnet, in general, is a graph where nodes are *synsets* (i.e., sets of synonyms) with semantic relations between these nodes. Each *synset* refers to a shared meaning. Semantic relations like hyponymy and meronymy are defined between synsets.

**Thesauri versus wordnet.** Although a wordnet can be seen and used as a thesaurus, it is much more (Miller et al., 1990). The main differences can be summarized as (i) a wordnet provides more accurate relations between words meanings, while in a thesaurus we can only know that two words are related, or that one word is more general/specific than another; (ii) a thesaurus does not define or designate the different meanings of its entries, while a wordnet does. This accuracy of a wordnet, over a thesaurus, enables building more effective and more accurate applications e.g., in medicine (Huang et al., 2009), text classification (Scott & Matwin, 1998), and others.

---

[1] We use "WordNet" to refer the Princeton wordnet, and "wordnet" to refer to resources with the same structure as WordNet.

Research on building ontologies for developing smart applications (e.g., semantic web, data integration, reasoning, knowledge management) has led to a deeper understanding of lexical semantics. For example, the OntoClean methodology (Guarino & Welty, 2009), upper level ontologies as DOLCE (Masolo et al., 2003) and BFO (Smith et al., 2015), gloss formulation (Jarrar, 2005, 2006), cross-lingual ontology alignment (Euzenat et al., 2005; Helou et al., 2016), the W3C Lemon model of exchanging and linking lexical resources (McCrae et al., 2011), and lexicon interoperability (Calzolari et al., 2013) are examples of semantic resources and tools that can help to better understand, validate, formalize and share lexical-semantic resources.

**WordNet versus Ontology.** Since WordNet has a similar structure as ontologies, it might be confused with an *ontology*. WordNet was developed with linguistic, rather than ontological analysis in mind as its authors explained: WordNet's criteria to judge whether a concept is sub/super of another concept is the rule "*If native speakers accept a sentence like: B is a kind of A*" (Miller et al., 1990). As argued in the next sections, defining and classifying concepts based on what native speakers perhaps naively accept may lead to inaccurate classification. This is because people generally have different beliefs; and most importantly, it is methodologically difficult to have access to people's minds (Smith, 2006) to validate what they intend and whether a classification really adheres to their beliefs or not. The following examples from WordNet 3.1 illustrate different kinds of cases, which at first glance look acceptable, but they are not, from an ontological viewpoint:

1. Stating that (Reflate$_2$ *Is-A* Inflate$_3$) and (Inflate$_3$ *Is-A* Change$_1$), and then stating (Reflate$_2$ *Is-A* Change$_1$) is logically useless. This is an implied relation, as it can be inferred from the first two hyponymy relations.

2. Saying (Islamic Calendar Month *Is-A* Month) sounds trivial but it is inaccurate, because Month is defined as one of the twelve divisions of the *calendar year*, which is a Gregorian year as defined in the gloss (i.e., an average month is about 30.43 days); however, an average of an Islamic month is 29.53 days. In fact, both months belong to two different calendar systems. A Gregorian month is 1/12 of a Gregorian year, while an Islamic year is a multiple of 12 lunar months.

3. Stating that Imaginary and Complex numbers are synonyms is inaccurate, as an Imaginary number is only a special case of a Complex number. Similarly, WordNet provides a poor classification of the types of numbers, e.g., Real, Rational, Natural, and Integer numbers are all subsumed by Number, while they subsume each other.

4. Defining Morning and Evening Stars as different stars is inaccurate. The Evening Star is defined as "A planet (usually Venus) seen at sunset in the western sky", and the Morning Star is defined as "A planet (usually Venus) seen just before sunrise in the eastern sky". That is, it is the same instance (i.e., Venus) that people see at different occasions.

5. Distinguishing between verbs and their verbal nouns in cases like the verb Learn "Gain knowledge or skills" and the noun Learning "The cognitive process of acquiring skill or knowledge", and similarly, Drink "Take in liquids" and Drinking "The act of consuming liquids", is ontologically confusing, since the notion of verb is a mere linguistic category. Ontologies capture the events that verbs denote rather than verbs themselves (see Subsection 4.3.3).

Gangemi et al. (2002) illustrated similar confusions between WordNet's concepts and individuals, between object and meta levels, in addition to violations of formal properties and heterogeneous levels of generality, and proposed to re-engineer and formalize the noun hierarchy to be a formal ontology. A topological analysis of the verb hierarchy was also discussed by Richens (2008) including anomalies, cycles, and unjustified symmetric and asymmetric rings; other hypernymy faults, called ring and isolator cases, are also identified by Liu et al. (2004).

As ontologies are strictly formal, specified in some logical language with formal semantics, it is possible to perform formal analysis and reasoning with them, which is obviously unsafe to do with wordnets as they lack formal semantics (Gangemi et al., 2010; Ceusters et al., 2005; Magnini et al., 2002). Additionally, instead of adhering to speakers' naïve beliefs, ontologies are intended to capture rigorous knowledge about a given world (Guarino, 1998), or according to Smith (2004), they represent entities in reality. In other words, the main difference between a wordnet and an ontology is not only *how* their content is structured or specified, but mainly about the level of accuracy and correctness of this content, as demonstrated above, which is called the level of *ontological precision* (Guarino, 2012) or *degree of formalization* (Magnini & Speranza, 2002). Kless et al. (2012) discussed the differences and similarities between ontologies and thesauri, and similarly, Hirst (2009) discussed how ontologies and lexicons can benefit from each other. Specialized thesauri and terminologies and their ontological use in the biomedical domain were also discussed by Ceusters et al. (2005), Smith et al. (2004), and Smith et al. (2006).

This article presents an Arabic wordnet built using ontological principles, called the *Arabic Ontology*. In addition to being formal, we also followed a strict methodology to validate our design choices. The definitions and classifications are validated and benchmarked against scientific advances and subject-matter experts' knowledge, as much as possible, instead of only adhering to people's naïve beliefs. As will be explained, our goal is to build a formal Arabic wordnet with as much ontological precision as possible. We conducted a coverage evaluation which shows that the top-levels of the ontology are able to top the majority of Arabic meanings – about 90% of the concepts used in the experiment were classified under nodes in the Arabic Ontology.

Although it is still ongoing work, the current version of the ontology is accessible through an online[2] lexicographic search engine (Jarrar & Amayreh, 2019) that also includes about 150 Arabic multilingual lexicons, which are being mapped and enriched using the ontology. The ontology, currently, is fully mapped to WordNet, Wikidata, and to about 13445 concepts in other Arabic lexicons. Correspondences between the top-levels of the Arabic Ontology and BFO, DOLCE, and DOLCE+DnS[3] are also discussed in Section 5. The ontology consists currently of about 1,300 well-investigated concepts in addition to 11,000 concepts that are partially validated and that we collected from specialized glossaries and other resources, and we are in the process of ontologizing them further. They are not fully ready, but we include them as we believe that they might be useful for some NLP tasks. Although the ontology includes several types of relations and formal expressions, in its current status, it is more of a *taxonomy*, rather than a richly axiomatized ontology.

This paper is structured as follows: Section 2 overviews related work. Section 3 presents the foundation of the Arabic Ontology and the methodology used to develop it. Section 4 provides an overview of the Arabic language and our related lexical and semantic choices. Section 5 presents the content of the ontology, and Section 6 evaluates its coverage. Section 7 illustrates the ontology portal and a framework for mapping it with 150 lexicons. Section 8 outlines our future directions.

**2. Related Work**

*2.1. Related Linguistic Ontologies*

The importance of developing linguistic ontologies is widely recognized with varying tendencies and for different purposes. As explained earlier, after developing the Princeton WordNet, there have been hundreds of wordnets developed for many languages with different coverage (see globalwordnet.org). For example, seven EU wordnets were built during the EuroWordNet project (Vossen, 1998); and similarly, the BalkaNet (Stamou et al., 2002) targeted Greek, Turkish, Romanian, Bulgarian, Czech and Serbian. Both EuroWordNet and BalkaNet were built following the same design as WordNet. Many efforts were also proposed to use wordnets as ontologies, so to facilitate multilingual knowledge access and processing (Magnini & Speranza, 2002). BabelNet (Navigli & Ponzetto, 2012), Yago (Suchanek et al., 2008), and ConceptNet (Speer et al., 2017) integrated wordnets with DBpedia and other resources into large knowledge graphs accessible through SPARQL endpoints. The OntoWordNet project (Gangemi et al., 2010) aimed at formalizing WordNet nouns to be a formal ontology by (re)interpreting synsets as classes or individuals, and the hypernymy links as subclass-of relations, in addition to checking the consistency of the overall results. Another promising attempt to develop a comprehensive linguistic ontology is called Ruthes. Although it is a Russian thesaurus, it is claimed to be a linguistic ontology (Loukachevitch & Dobrov, 2014). The authors differentiate it from Russian wordnets, and stressed the difference between a wordnet synset and an ontological concept, which they consider to be a unit of thought.

---

[2] http://ontology.birzeit.edu/concept/293198
[3] http://ontologydesignpatterns.org/wiki/Ontology:DOLCE+DnS_Ultralite

There are other linguistically-motivated ontologies aiming to cover some domains or aspects of a language. For example, a linguistic ontology focusing on spatial expressions for natural language processing is proposed by Bateman et al. (2010), and a linguistic ontology of events is proposed by Dini and Bertinetto (1995). A related work called VerbNet was proposed to classify English (Kipper et al., 2006) and Arabic (Mousser, 2010) verbs. Verbs sharing the same class have the same syntactic descriptions and argument structure. Although this classification of verbs is not an ontological classification, as it is not based on verbs semantics, it was recently connected with other semantic and ontological resources with the Framester framework (Gangemi et al., 2016). A small upper linguistic ontology for English, German, and Dutch, called the Generalized Upper Model, was suggested by Bateman et al. (1995), which is claimed to be a task and domain independent ontology for language expressions. Its main goal is to interface surface linguistic realizations with conceptual representations. The need for such an interface ontology was acknowledged by Ceusters et al. (1998) who also argued that it should be close enough to language realizations. Another type of ontology to describe linguistic descriptions is GOLD (Farrar & Langendoen, 2003), but this ontology does not aim to capture the semantics of terms. It mainly classifies morphological notations, such as expressions, grammar, and meta-concepts.

*2.2. Related Arabic Resources*

**The Arabic WordNet** (Elkateb et al., 2006) consists of about 10,000 synsets and is a literally translated small subset of WordNet. It was constructed by first identifying a set of synsets in WordNet, called the Common Base Concepts, and then manually translating them into Arabic. These base concepts exist in 12 languages (in EuroWordNet and BalkaNet, thus they are assumed to also exist in Arabic (Elkateb et al., 2006). They were extended mostly downwards with more specific concepts, and upwards with more general concepts, to improve the maximal connectivity of those base concepts (Elkateb et al., 2006). That is, the Arabic WordNet is a subset of WordNet. The semantic relations between the translated synsets were copied as is, from WordNet. As WordNet was mapped to the SUMO ontology (Niles & Pease, 2001), all mapped concepts were also included in the Arabic WordNet. Compared with the Arabic Ontology, the Arabic WordNet inherits all logical and ontological concerns from WordNet since it is a literal translation and subset of it, while the Arabic Ontology is formal and avoids such concerns. Additionally, concepts in the Arabic Ontology are natively defined and classified, rather than translated from another language. Nevertheless, all concepts in the ontology are mapped to WordNet and thus to the Arabic WordNet, as will be discussed later.

**Other Arabic resources**. There are other small linguistic-oriented Arabic ontologies developed for particular use, e.g., for news items (Abdulhussein & Raisan, 2015), biology (Al Azemi & Al-Radaideh, 2011), ontology extraction (Al Zamil et al., 2014, Mazari et al., 2012) among other domains and tasks. Kamal et al. (2015) experimented the idea of generating an Arabic ontology automatically, by linking the MERIT multilingual database with WordNet and with SUMO. The generated ontology in this experiment is not accessible, but it is obvious that such an approach would generate only a thesaurus, rather than a wordnet or an ontology. The term "Arabic ontology" was used by Belkredim and El-Sebai (2009) and by Aliane et al. (2010) for representing morphological categories in Arabic in an ontological manner, which is not related to our use of this term.

## 3. Foundations of the Arabic Ontology

*3.1. Arabic Ontology Structure*

The general structure (i.e. core data model) of the Arabic Ontology is similar to the structure of WordNet, as depicted in Figure 1. Maintaining a similar data model helps in concept-mapping and interoperability with wordnet resources. Each concept in the ontology is given a *unique concept identifier* (URI), informally described by a *gloss*, and lexicalized by one or more of synonymous lemma terms. Each term-concept pair is called a *sense* and is given a *SenseID*. A set of senses is called *synset*. As will be described in Section 4, concepts and senses are described by further attributes such as *era* and *area* – to specify when and where it is used, lexicalization type (e.g., classic, dialect, technical, etc.), example sentence, example instances, ontological analysis, and others. *Semantic relations* (e.g., *SubTypeOf, PartOf*, and others) are defined between concepts. We additionally allow some important *individuals* to be included in the ontology, such as individual countries and seas. These individuals are lexicalized in the same way as concepts, however they are given separate *ID*s, and linked with their concepts through the *InstanceOf* relation. The Arabic Ontology includes also formal axioms, which are currently placed in the ontological analysis profile, as will be explained later.

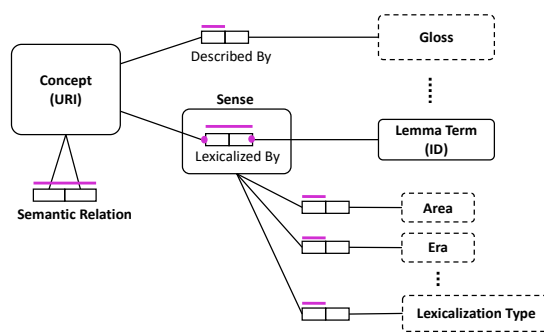

**Figure 1.** The core metamodel of the Arabic Ontology, formalized using the ORM notation (Hodrob et al., 2010).

*3.2. The Notion of Concept*

The Arabic Ontology is a concept-centered view of the knowledge that the Arabic terms convey. As an ontology, concepts are *classes of individuals* in the formal set-theoretic sense. Organizing knowledge in this extensional concept-centered manner enables IT applications to communicate meaningfully and reason about such knowledge when equipped with appropriate software. This is unlike wordnets, which do not define concepts, but rather, define synsets that "*signify* that the concepts exist" (Miller et al., 1990). In other words, concepts in the ontology are the main building blocks, and terms are merely lexicalizations (i.e., names) of concepts.

The term concept is widely used among the lexical semantics community, and its definition was standardized by ISO TC 37 as a "*unit of knowledge created by a unique combination of characteristics*" (ISO, 2000). In the same standard, objects (e.g., 'Saturn', 'the Eiffel Tower') are also called *individual concepts*. As stated by Smith et al. (2004), the ISO definition is based on Eugen Wüster's work (Wüster, 2003) who argued that concepts and individual objects are both *thoughts* existing in our minds. This definition was criticized by Smith (2004), Smith et al. (2004), and Smith (2006), that constructing concepts as "thoughts" does not help us to benchmark the correctness of our concept system – as we cannot gain access to the interiors of each other's brains.

In this paper, we use the term *concept* (in Arabic *mafhūm* مَفْهُوم) as it is the common term in the ontology and in the AI and logic communities (Baader et al., 2007). However, our notion of concept is different from the ISO's notion of concept, which *allows concepts not to be instantiated.* A concept in the Arabic Ontology is anything that is instantiated, and that signifies the characteristics that its instances have in common, whether these instances exist in reality or not. There are no concepts in the ontology that cannot have instances, and no instance can be a concept at the same time. As stated earlier, we do not restrict every instance of a concept to have a location in time and space, as we need to allow abstract and social entities in the ontology.

The notion of instance is called *māṣadaq* (ماصَدق), or *miṣdāq* (مِصْداق) plural *maṣādīq* (مَصَاديق) in the old Arabic philosophy literature (e.g., Ibn Mohammad, 1983). Some researchers distinguish, correctly, between an instance and a particular (*fard* "فَرد"), to say e.g., my cat (which I call Foxi) is a particular, and that this particular is an instance of the concept cat. In this paper, we use the terms instance, particular, or individual to mean the same thing – for the sake of simplicity.

**Formal Semantics**

A concept can be extensionally seen in the typical formal way as a set of individuals, and so the relations between them. However, it should not be interpreted in the way described by Genesereth and Nilsson (1987), which refers to a particular state of affairs: given a concept *c*, its formal extensional interpretation $c^I$ is a subset of domain $D$: $c^I \subseteq D$. Instead, concepts are interpreted *intensionally* as defined by Guarino (1998), which refers to "admissible worlds", or called a domain space <D, W>, rather than a domain D. In what follows we adapt this definition for our purposes.

**Definition (Formal Interpretation)**

Given a concept *c*, its intensional interpretation $c^I$ is defined on a *domain space* <D, W> as a function $c^I: W \rightarrow 2^D$, where D is a domain and W is a set of maximal states of affairs on D. For a concept *c*, the set $E_c = \{c^I(w) \mid w \in W\}$ is the set of the admissible extensions of *c*. Two concepts having the same set of *admissible* instances, in all states of affairs, are considered the same concept.

*3.3. Semantic Relations*

Any semantic relationship is allowed in the Arabic Ontology. The subsumption relationship is given a greater attention and is considered the backbone of the Arabic Ontology, because of its importance in characterizing and distinguishing between concepts (Guarino, 98). This relation is analogous to, but stricter than, the hyponym relation in WordNet. We use the definition of *subsumption* as defined by Guarino and Welty (2009), which says that the concept $c_1$ subsumes $c_2$, iff every instance of $c_2$ is an instance of $c_1$, in every possible state of affairs. The ontology, in its current status, is more of a taxonomy with disjoint concepts, rather than a richly axiomatized ontology. This does not mean that formal class axioms and expressions are not allowed; but since the ontology will be mainly used for NLP tasks, rather than formal reasoning, we chose to place formal class axioms and expressions separately in concept *profiles* (see Section 7). The parthood relationship is not fully formalized in the current version of the Arabic Ontology. Our progress in formalizing related relationships (e.g., *occurrent-part-of, temporal-part-of*, *isotypic-part-of*, *homeomeric-in*, *cumulative-with*, and *is-telic-in*) can be found in a previous work (Jarrar & Ceusters, 2017).

*3.4. Benchmarking and Ontological Analysis*

A major challenge we faced while developing the ontology was what the ontology should capture and adhere to, and on what basis the correctness of the ontology content can be benchmarked. For example, should concepts be defined and classified based on what Arabic speakers commonly believe, should we adopt a certain lexicon and formalize it, should we rely on what the scientific literature accepts, or should we build the ontology based on what we, the ontology builders, believe. The benchmarking methodology is important in order to evaluate whether the ontology contains inconsistent or incorrect facts, and to validate our choices while building the ontology, especially as many ontology developers are involved in the process.

Developing an ontology based on what Arabic speakers commonly believe or based on a certain lexicon may lead to poor semantics, false beliefs, inconsistent, and/or vague classifications. Similarly, developing an ontology based only on scientifically-verified knowledge is difficult, because there are concepts that do not have scientific theories underpinning them, such as social conventions.

The methodology we decided to follow for benchmarking consists of three levels of benchmarking preferences, as depicted in Figure 2. That is, we benchmark the ontological precision of our concept definitions and classifications against the following, in order:

1. Scientific knowledge, which scientists typically accept on the basis of experimentation and verification and commonly agree about. If no mature answer is found in the state-of-art scientific discoveries, then against,
2. Subject-matter experts' and domain knowledge and conventions. If no answer can be synthesized or attained from experts' knowledge, then against,
3. Commonsense knowledge that can be repeatedly found in quality lexicons and literature.

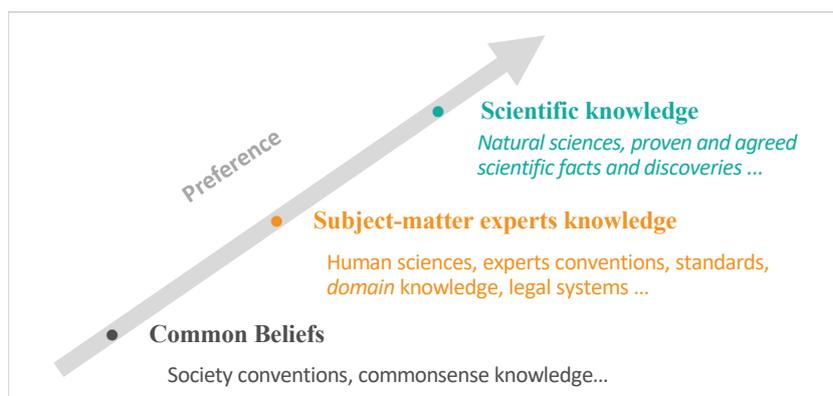

**Figure 2.** Preferred levels of ontology benchmarking.

In this way, we aim to achieve a high ontological precision as much as possible, and try to avoid building an ontology that relies solely on speakers' or one's beliefs.

For example, when we defined and classified organisms, we had to investigate the scientific literature and to rely on scientific assertions, and thus the correctness of our classification of organisms is benchmarked against the advances in Biology. Unlike WordNet and most lexicons, we did not define a virus, for instance, as a type of organism because it cannot reproduce itself independently, which is the main characteristic of organisms. Similarly, we had to consult Astronomy references for classifying astronomical bodies (stars, planets, satellites, etc.), and Geography for classifying the types of lands and water regions (oceans, seas, lakes, rivers, etc.). However, when defining the types of states and government systems, which do not have a scientific law underpinning them, we had to rely on law and political sciences to understand and classify them, i.e., domain and subject-matter experts' knowledge.

In such a case, and as government systems are largely influenced by different legal systems and conventions, our classification of states and government systems tries to synthesize and harmonize disparities of this domain knowledge. Finally, there are cases where neither science nor subject-matter experts can help us; for example, when classifying the types of love and hate, the types of sites, buildings and shelters, and others. In such cases, which are typically social conventions, we had to investigate and rely on the commonsense knowledge found in good Arabic lexicons and appropriate literature.

In addition to the above three levels of benchmarking preferences, we also used well-founded ontological analysis, using the OntoClean methodology (Guarino & Welty, 2009), for all decisions made, and we documented these decisions. For each concept, we introduce an *ontological analysis profile,*[4] in which we identify and document: (*i*) what characteristics distinguish the instances of this concept from the instances of the sibling concepts, (*ii*) example instances, (*iii*) what identity criteria instances have and/or inherit, (*iv*) whether a concept is rigid or not, and (*v*) formal axioms and expressions, among others. To avoid possible overlapping and confusion between concepts and the identity criteria of their instances, the subsumption relations in the Arabic Ontology form a tree (i.e., multiple inheritance is not allowed), and all sibling concepts are mutually disjoint. Each concept has a distinct set of distinguishing characteristics, and when a concept is specialized into sub-concept(s), each sub-concept adds further characteristics to those carried by the super-concept. As will be discussed in Section 5, the top levels of the ontology provide a further methodological framework, for developing and governing the logical and ontological quality of the lower levels.

*3.5. Gloss Formulation Guidelines*

In addition to its formal definition, we propose each concept in the ontology to be informally defined by a *gloss*. The purpose of a gloss (see Jarrar, 2006) is not to provide or catalogue general information and comments as in encyclopedias or as used in `owl:comment`, but rather, to state the critical and distinguishing characteristics that all instances of a concept have in common, in an informal but controlled way – as the following:

**Step 1: Start with the supertype** of the concept being defined. For example, "Object: An entity that…", "Physical Object: An object that ...", "Social Object: An object…". That is, the first term in the gloss of every concept is its direct supertype in the ontology.

**Step 2: List only the most distinguishing and intrinsic characteristics that specialize** the concept from its supertype, and that differentiate it from other concepts in the same level. For example, "*Object: An entity that is wholly and independently present in time, and is recognized either for its concrete or social existence*", and "*Physical object: An object that occupies space, and is recognized by senses or measuring tools*". Notice that we do not need to repeat that the physical object is present in time, as that is a characteristic of its supertype.[5]

**Step 3: Write the distinguishing characteristics in the form of a sequence of propositions** to help the reader to easily mentally rebuild the concept being defined in a declarative and non-narrative manner. For example, compare the following two glosses of *social object*: (1) "*An object is social if it can be understood and recognized by people in a social system that exists; social objects are also those can be represented by physical objects*", and (2) "*An object that is recognized for its social existence, and can be represented by physical objects*". The second is formed as propositions, while the first is more narrative.

---

[4] The `<Ontological Analysis>` field can be accessed online by adding the `/profile` to the concept URI, such as http://ontology.birzeit.edu/concept/293200/profile , or http://Ontology.birzeit.edu/concept/334000112/profile

[5] It is worth noting that step 1 and step 2 are consistent with what is called *āl-ḥadd* (التعريف بالحدّ والرسم) defined by al-Ġazālīy (1058-1111 AD) in his book *āl-ḥodūd* (الحدود) as described by Alaasam (1997).

In this manner, a gloss provides a concise documentation of characteristics. We have noticed that this way of formulating glosses provided *methodological support* for junior ontology developers as it guides them to find and "organize" distinguishing characteristics. Furthermore, people who are not interested to see the formal definitions and classification of concepts can still use the ontology as an informal lexicon.

## 4. The Arabic Lexical Level

*4.1. Types of Arabic*

Arabic is an official language in 23 countries, spoken by more than 300 million people, a liturgical language for 1.6 billion Muslims, and is one of the six official languages of the United Nations. Arabic has multiple forms: Classical Arabic, Modern Standard Arabic (MSA), and Dialectal Arabic (see Jarrar et al., 2014; Jarrar et al., 2016). Classical Arabic is the Arabic used between the 6$^{th}$ and the 15$^{th}$ centuries. Modern Standard Arabic is the Arabic that all Arabs understand, and used for formal communication including news, media, education, and literature. Dialectal Arabic is the form of Arabic that is used in the informal day-to-day communication, such as in conversations and chatting. Additionally, and like languages that are widely spoken in many different countries and cultures (Kramsch 1998), there are many terms and concepts in Arabic that are known only in specific regions, especially those related to food, religion, and local culture. Also, the same concepts might be lexicalized by different terms in different regions, and the same terms might be used for completely, or partially, different concepts. In addition, there is a huge amount of Arabic literature from the middle-ages that is well documented and accessible in modern digital libraries, but which contains obsolete terms and obsolete concepts. Although all types of Arabic are enabled in the Arabic Ontology, we aim primarily at capturing the Modern Standard Arabic (MSA) that is typically found in MSA lexicons.

*4.2. Introducing the Spatiotemporal Dimension for term appropriateness*

Although we primarily target MSA, we believe that other forms of Arabic should also be supported. Therefore, we propose to annotate both concepts and senses (i.e., term-concepts) with additional spatiotemporal information (Figure 1).

**First**: we propose each concept to be annotated with *area* and *era* attributes to encode in which region(s) and which period a concept is or was used. The concepts that we consider as MSA are typically annotated with the `MostArabCountries` area and the `Modern` era (see Figure 3). Other concepts, e.g., the concept that is called *hayūliyy* (هَيُولِيّ) is annotated with `mid-ages` era, and the food *ṣamṣat* (صمصة) is annotated with the `NorthAfrica` area.

**Second:** we propose each term-concept (i.e., sense) to be annotated with era and area, as well as with a *lexicalization type,* such as `dialectal-name, scientific-name,` or `legal-name.` For example, and as illustrated in Figure 3, the concept of *heartburn*, which is called *ḥumūḍat* (حُموضَة) or *ḥurqat* (حُرْقَة) in MSA, is also informally called *ḥazaz* (حَزاز) in Palestine and Jordan. Thus, instead of mixing *ḥazaz* (حَزاز) with the other MSA terms in the same synset, the lexicalization type of this term-concept is marked with `dialectal-name,` in the `Palestine&Jordan` area.

```xml
<concept conceptID=":293198" area=":MostArabCountries" era=":Modern">
 ...
</concept>
<Concept conceptID=":291234" area=":MostArabCountries" era=":Modern">
 <Synset>
  <Sense ID=":26754" Term="حُموضة" area=":MostArabCountries" era=":Modern" lexicalizationType=":MSA"/>
  <Sense ID=":26747" Term="حَرْقة" area=":MostArabCountries" era=":Modern" lexicalizationType=":MSA"/>
  <Sense ID=":26249" Term="حَزاز" area=":Palestine&Jordan" era=":Modern" lexicalizationType=":DA"/>
  ...
```

```
<Concept conceptID=":50856" area=":MostArabCountries" era=":Modern">
 <Synset>
  <Sense ID=":26754" Term="عزبة" area=":Egypt" era=":Modern" lexicalizationType=":MSA"/>
  <Sense ID=":26747" Term="ضيعة" area=":Lebanon" era=":Modern" lexicalizationType=":MSA"/>
  . . .
```

**Figure 3.** Illustration of geospatial and temporal annotations.

*4.3. The Morpho-Semantic Level*

This section discusses how the semantic and morphological levels interact, and the implication on the ontology.

*4.3.1. Inflectional Morphology (Lexeme and Lemma)*

Arabic is among the highly inflected languages (Ryding, 2014; Boudelaa & Marslen-Wilson, 2004). Thousands of words can be generated from one stem by attaching affixes and clitics to it. There are many types of clitics and many types of affixes in Arabic (to indicate e.g., gender, number, mood, tense, case, voice, etc.). One may also attach more than one prefix and more than one suffix to a stem at the same time. For example, from the stem *katab,* which means "write", one may generate the word *wasayaktubwnahā* (وَسَيَكْتُبونَها) to mean "and they will write it". English, in comparison, has a smaller set of inflection features, e.g., the inflections of *write* are only *writes*, *wrote*, *written*, and *writing*. The set of all words that are inflections of each other is called *lexeme*. Since inflectional morphology does not modify the core meaning, all words in a lexeme have the same core meaning (Ryding, 2014). One of the words in a lexeme can be selected to conventionally represent the lexeme, called *lemma*. Lemmas are typically used as *headwords* in dictionaries. For a noun lexeme in Arabic the lemma is typically selected to be the singular masculine form; while for a verb lexeme, it is the 3rd person, singular perfective form.

We adopt the notion of lexeme and the notion of lemma in the ontology. We restrict synsets to contain only lemmas, which represent all inflections in their lexeme. Thus, there is no need to include other inflections in the synset. We try to use, as much as possible, the lemmas found in the LDC's SAMA 3.1 database, which is the database of the Standard Arabic Morphological Analyzer (Maamouri et al., 2010). As lemmas in SAMA are used and linked with many other lexical resources and corpora, this maximizes the extendibility and compatibility of the ontology with other resources.

*4.3.2. Derivational Morphology (Roots and Patterns)*

Derivational morphology is another important way of generating words in Arabic (Boudelaa & Marslen-Wilson, 2004; Ryding, 2014; Elkateb et al., 2006). Arabic words are derived from *roots* using a rich set of *patterns*. The root-pattern derivation exists in other languages (e.g., the English -er and -ing with *write,* to generate *writer* and *writing*), but Arabic is highly derivational compared with other languages (Boudelaa et al., 2010). The Lexicon of Modern Arabic contains only 5,778 roots, from which 32,300 lemmas in the lexicon are derived. For example, given the root *ktb*, we may derive the verb *kataba* (write), *kitāba* (the process of writing, or called verbal noun), *kātib* (the writer or author, the agent who did the writing, which is called *active participle*), *maktūb* (writing, what was written, which is called *passive participle*), *kitāb* (book or letter, which is a kind of writing), *kutayyib* (booklet, a small-size book, or called diminutive noun), *kātiba* (typewriter, which is called the *instrument noun* of writing), *maktaba* (library), *maktab* (desk or office, where the process of writing happens, or called the *noun place*), to name a few.

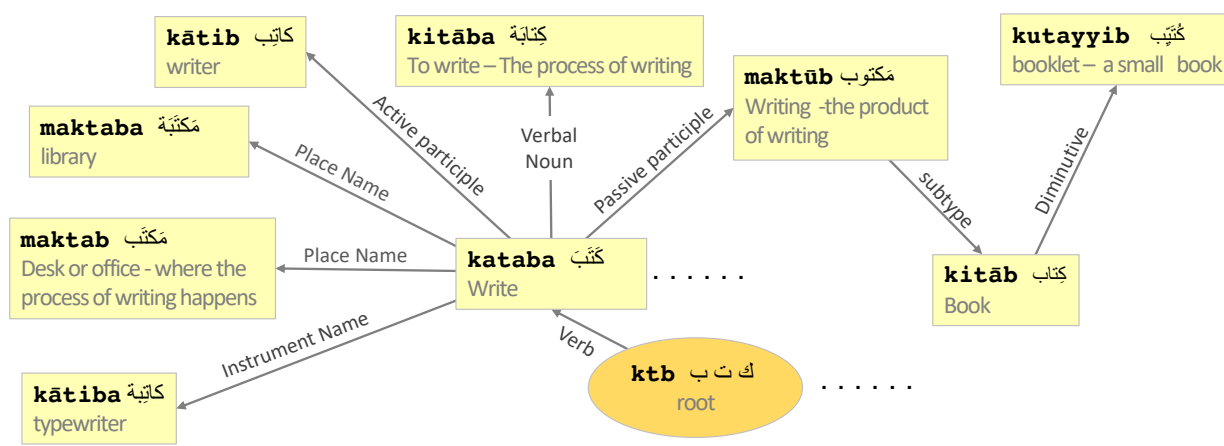

**Figure 4.** Example of Arabic Morpho-Semantic Relations

Roots in Arabic are not words in the language, but rather, sequences of letters carrying the general (or the *source)* meaning that all derivations, and their inflections, share or have in common (Ryding, 2014; Boudelaa et al., 2004, 2010). Lexicons use roots as primitive units of meaning or as semantic fields to organize and index their entries. Since derivations are derived using a set of predefined patterns, all derivations are predictable in form, and all are semantically related with each other, as illustrated in Figure 4. Some researchers also believe that roots and patterns can be used to infer the semantics of their derivatives (Elkateb et al., 2006).

Incorporating roots and patterns in the ontology is considered future work. Although the idea of ontologizing derivations and automatically inferring the meaning of derivatives sounds promising, we believe this topic needs more investigation, especially specifying the notion of semantics that roots denote and formalization of derivations.

*4.3.3. Capturing verbal nouns instead of verb synsets*

Every verb in Arabic has a *maṣdar*, which is the name (i.e., a noun word) of the event that the verb refers to. For example, the maṣdar of the verb *kataba* (write) is *kitaba* ("writing" or "to write"), and the maṣdar of *ʾakala* (eat) is *ʾakel* ("eating" or "to eat"). The notion of maṣdar is close to the notion of *verbal noun* or *gerund* in other languages, such as verb+ing in English. There is no ontological difference between a verb and its verbal noun. For example, we may say "John wrote his thesis in six months" or "John's writing of his thesis took 6 months", and in both statements, we refer to the same event of writing. Different languages may provide different verbs' time preferences (i.e., tenses) but this is ontologically irrelevant because ontologies capture instances of events rather than the way we refer to them (Jarrar & Ceusters, 2017).

The notion of verb itself is a linguistic category and has no ontological significance – as there are no ontological categories corresponding to verbs. Ontologies capture primarily the events or processes that verbs denote, but not verbs themselves. Thus, instead of capturing verbs in the ontology we capture their maṣdars (verbal nouns). In other words, there are no verb synsets in the Arabic Ontology. The semantics of a verb can be understood through its maṣdar and the events that this maṣdar lexicalizes. It is also important to remark that although the majority of the verbs denote events, there are some *stative verbs,* which should be classified under *state* in the ontology, e.g., "It *costs* 20$", "I *know* him", or "I *need* help".

*4.4. Synonymy is an equivalence relation*

WordNet defines synonymy as follows: "two expressions are synonymous in a linguistic context C if the substitution of one for the other in C does not alter the truth value" (Miller et al., 1990). This definition is problematic: first, substitutability is not a precise measure. One may say that e.g., "The woman I met yesterday was very smart" and "The person I met yesterday was very smart". Substituting person and woman did not alter the truth value in these two statements. That is, it is likely that when substituting a word with another, it's a supertype of the other. Second, and most importantly, ontologies do not represent knowledge based on linguistic contexts, but rather, extensions of concepts. Therefore, we define synonymy between two terms, if and only if, both terms lexicalize the same concept.

**Definition (Synonymy Relation)**

Given two terms $t_1$ and $t_2$ lexicalizing concepts $c_1$ and $c_2$, respectively, then $t_1$ and $t_2$ are considered to be synonymous *iff* $c_1 = c_2$. In this way, synonymy can be defined as an equivalence relation $=_c$ between terms lexicalizing the same concept, thus it is a reflexive, symmetric and transitive relation.

*4.5. Polysemy*

A term that is used to lexicalize more than one concept is called polysemous. We do not introduce any special treatment of polysemous terms. Like in wordnets, if a term appears in more than one synset, it is polysemous. At this early phase, it is difficult to quantify or estimate the size of polysemy in the current version of the ontology, as only a small set of terms and concepts are captured.

*4.6. Concept lexicalization and lexical gaps*

One of the differences between application ontologies and linguistic ontologies is that linguistic ontologies are mainly concerned with capturing *named concepts.* Named (also called lexicalized) concepts (Hirst, 2009; Miller et al., 1990) are those concepts that are expressed either by one word in a language, or by multiple words that are repeatedly used among speakers. For example, "infectious disease", "medical examination", "public health organization", or "developing country" are examples of multiple-word expressions that we typically find as entries in general or domain-specific lexicons. On the other hand, application ontologies may capture ad hoc concepts and their names can be free expressions. For example, one of the concepts in the Gene ontology is named "*Regulation of voltage-gated potassium channel activity involved in ventricular cardiac muscle cell action potential repolarization*", which are not typical to find in a linguistic ontology as it is not widely used.

The Arabic Ontology aims at capturing the concepts that have single- or multiple-word lexicalizations in the Arabic language. This means that concepts and their lexicalizations in the ontology are restricted to the common and repeatedly used concepts. To adhere to this restriction, we use a set of Arabic lexicons (see Subsection 7.2), such that each lexicalization in the ontology is a lexical entry in one or many lexicons. This does not mean that domain-specific concepts cannot be part of the ontology if they are communicated within a domain.

*Remark on lexical gaps*: A lexical gap simply is the lacking of a lexicalization of some concepts, especially relative to other languages (Hirst, 2009). Although the Arabic Ontology is a monolingual ontology, when classifying concepts, sometimes we need to introduce a new concept and a new lexicalization, in order to better organize the ontological hierarchy. We allow such new concepts and lexicalizations only if they are highly important for maintaining the intuitiveness and the organization of the ontology hierarchy. At this stage, in which the ontology consists of about 1,300 concepts, there are only 4 concepts that we had to exceptionally introduce. For example, when classifying the types of lands, which are 37 types in the ontology, mostly distinguished based on their topography, we had the choice of either have them all under one concept, which would make it difficult to describe them, or classify them under three types, which would be called in English: flat, coarse, and non-coarse land. Although flat and coarse lands are lexicalized concepts in Arabic, there is no named concept in Arabic equivalent to non-coarse land, which was important to introduce for organization purposes.

## 5. Overview of the Arabic Ontology

This section gives an overview of the ontology in a top-down manner. After developing the top levels of the ontology and collecting many low-level concepts, we iterate to bridge between both levels. As explained in Subsection 7.27.2, we collected and digitized many Arabic resources (about 150 lexicons) including old, typical and monolingual lexicons as well as modern, multilingual, and domain-specific glossaries and thesauri. The idea is to have enough rich lexical semantic resources to query and find, for example, all terms related to *time* and their definitions, all synonyms related to *organisms* in multiple languages and their definitions, and so forth. Such sets of terms are then used as *candidate* concepts (i.e., as an initial step), to be revised, grouped and linked with the top levels, and to keep improving their definition and classification and benchmark them as discussed in Section 3.4.

Every concept is classified under (i.e., derived from) the top-level concepts. That is, the top-level concepts are used as a foundation for the more specific concepts; thus, they are used as part of the ontology itself, rather than using them as semantic fields or separate top concepts – as SUMO was used to top WordNet (Elkateb et al., 2006). Using the top-level concepts in this way is important for two reasons. *First*, they help **to avoid logical and ontological mistakes** at the lower levels, especially if one tries to define a concept under two disjoint concepts, e.g., *state* under both *Geopolitical Area* and *Government.* Such cases of confusing meanings can be detected using top levels. *Second*, they provide a **methodological framework** to assist in defining the lower levels. Instead of building the ontology in a bottom-up manner (e.g., defining a common concept of two specific concepts, which is not only difficult but also prone to error), it is easier to specialize the lower levels and derive them from the top levels. The top levels play the role of a methodological tool to analyze and distinguish between the lower levels.

Although our work was inspired by the work in upper ontologies (especially BFO, DOLCE, DOLCE+DnS, and SUMO), and although we reused some notions from these ontologies, the top-levels of the Arabic ontology are not claimed to be language independent or generic categories. Our goal was only to develop a comprehensive set of Arabic-specific categories of most meanings of Arabic words. Concepts in those upper ontologies are not necessarily *named concepts*, thus they can be *silent concepts* (i.e., invented by the authors of those ontologies). This is not the case for the Arabic Ontology, as every concept in our top levels is a named concept in Arabic.

Figure 5 illustrates some top levels of the ontology. The ontology is lexicalized in Arabic, but in this paper, we provide *an approximated English translation* – only for the sake of scientific publication and to compare it with related work. In addition, each concept is lexicalized by a synset (i.e., one or multiple terms), but in this paper we show only one term, for the sake simplicity. The following subsections give an overview of each branch of the ontology, but one may access the online version for the full ontological analysis, synsets, glosses, formal axioms, example instances, and the mappings to WordNet and correspondences to BFO, DOLCE, and DOLCE+DnS.

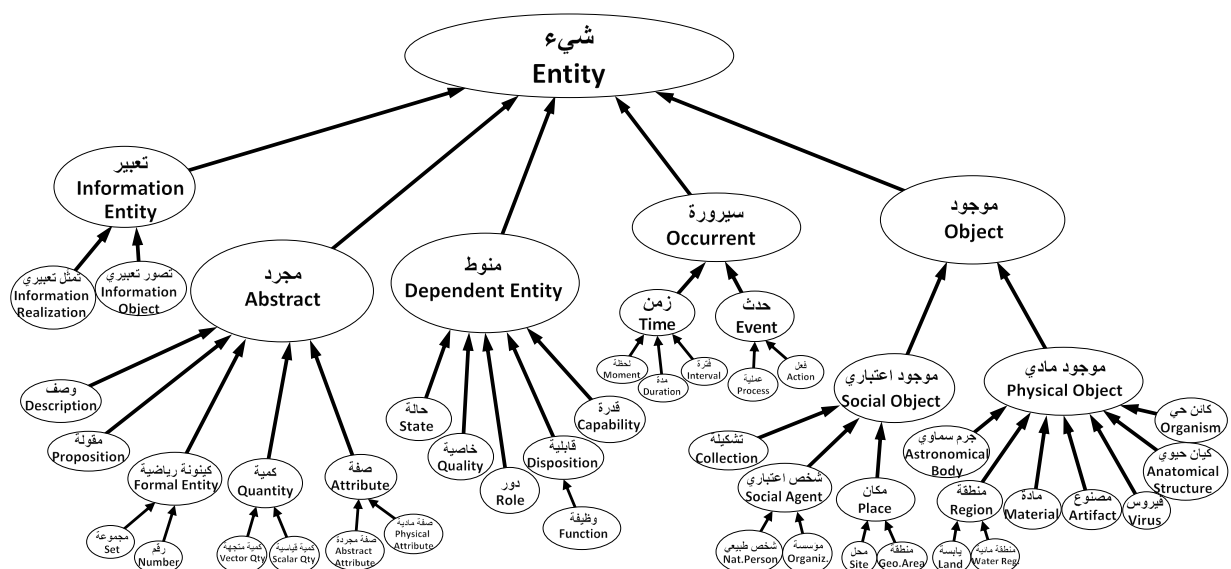

**Figure 5.** Top three levels of the Arabic Ontology.

*5.1. Object {موجود}*

An object is wholly and independently present in time, maintaining its identity, and will always be recognized as itself during its entire existence. This corresponds to independent continuant in BFO, endurant in DOLCE, object in DOLCE+DnS, and to the union of objects and physical systems in SUMO.

*Physical objects* {موجود مادي} are located in space and time, bear physical qualities that give them their physical properties, like mass and height. They correspond to material entities in BFO, physical objects in WordNet and DOLCE+DnS, and to physical endurants in DOLCE. We currently have about 600 subtypes of physical objects: 250 *organisms* (e.g., animal, plant, eubacteria); 130 *anatomical structures* (e.g., organ, cell, tissue, cranial cavity); 6 types of *virus* – unlike most ontologies we do not classify viruses under organisms because they cannot reproduce themselves independently, as organisms do; 150 *geographical regions* (e.g., land and water regions) which are different from the notion of place that is a social object; 15 *astronomical bodies* that exist naturally in space (like star, satellite, and comet). The planet Earth itself is an astronomical body, and every region in it is a geographic area. The *material* branch is not elaborated yet, but it corresponds to material in WordNet and to substance in SUMO and DOLCE+DnS. The notion of *artifact* is not well studied yet and we are hesitating to include it in the Arabic Ontology – although it is introduced in DOLCE and SUMO, because distinguishing what is an artifact is not clear (i.e., intentionally vs. naturally and fully vs. partially made) specially as we do not allow multiple inheritance and multiple classification of instances.

*Social objects* {موجود اعتباري} are realized for their social existence and represented by physical objects. They themselves do not physically exist, but they are socially realized and agreed upon between people or recognized by administration. This concept corresponds to *social objects* in DOLCE and DOLCE+DnS and largely to immaterial entities in BFO. We currently have about 200 subtypes of social objects, most of them are under *place*. A place is a social object that conveys a physical geographical region, but is realized by the type of activities carried out in it, or by the physical objects occupying it. This is not the same as physical places (which we call *geographical regions*, described above), which are physical entities. Examples of social places are *geopolitical areas* (like country, county, city, or neighborhood) or *sites* (like market, plant, road, square, shelter, or house). This notion of place corresponds to non-physical place in DOLCE (or called place in DOLCE+DnS) and to some extent to site in BFO. *Collections* and *social* *agents* are also social objects but they are not much elaborated yet. A collection (e.g., the Barcelona football team, and the band Queen) consists of a plurality of physical objects realized as a whole. This corresponds to collection in DOLCE+DnS, SUMO and WordNet, to object aggregate in BFO, and to agentive social object in DOLCE. A social agent, on the other hand, is a social object whose identity is realized and gained by its capability of performing actions related to this identity, such as committees, corporations, and foundations like the Futbol Club Barcelona, the Google Inc, or the Buffalo General Hospital. Google Inc, the legal entity is a social agent before and independently of being located or embodied in a building and before anyone is employed in it. The purpose, role, obligations, and responsibilities of a social agent should be known within or recognized by DOLCE.[6]

*5.2. Occurrent {سيرورة}*

As in BFO, an occurrent is an entity that happens in time and every part of it is an occurrent. This corresponds also to perdurants in DOLCE except for states and times. Times (also called temporal regions), which are occurrents in BFO, are abstracts in DOLCE. States, which we classified under dependent entity, are perdurants in DOLCE. We classify occurrents into events and time.

*Events* {حدث} exist in time by happening, have temporal proper parts, and depend on a physical entity to happen, like processes in BFO. Examples of events are sleeping and living. They correspond relatively to the union of events and physical processes in WordNet, to processes in SUMO, and to the union of DOLCE's processes, accomplishments, and achievements. We currently have two subtypes of events: actions and processes. Actions are those non-accumulative events that have peak moments (e.g., hitting and knocking), and processes are those cumulative events (i.e., the sum of two events is also an instance of the same event, such as growth and accumulation (Jarrar & Ceusters, 2017). As explained at the end of Section 4.3, we have no verb synsets in the ontology. Instead of capturing verbs we capture their maṣdar (i.e., verbal nouns) that refer to processes, actions, or events, like the process of sleeping and the process of growing, rather than verbs sleep and grow. We have collected most Arabic lemmas that are verbal nouns (~15k lexical entries), and classified them based on their semantic root. We use these lemmas as candidates for types of events.

*Time* {زمن | وقت} is a region in the timeline, or a portion of time, similar to *temporal region* in BFO and DOLCE. We currently have about 200 subtypes of time classified under three subtypes: 100 intervals, 70 durations, and 30 moments. An *interval* (e.g., year, week, day, morning, evening, hour, second) is a portion of time with starting and ending points, its length is based on the temporal dimension of some astronomical events. A *duration* is also an amount of time, but its length depends on the life (i.e., the temporal dimension) of certain objects or non-astrological events, such as age, era, youth, period, stage, phase, cycle, and season. A *moment* is an amount of time with fiat boundaries – it is conventionally considered as one point of time, such as now, the end of his career, the start of a landing, a sunrise, or a menopause.

*Dependent Entity* {منوط} whose existence depends on the existence of other entities, is similar to BFO's specifically dependent entities. We currently have five subtypes: role, disposition, capability, state, and quality; which we did not extend further yet. Roles and dispositions correspond to those in BFO. Roles describe the activities carried out by objects to fulfill a certain purpose (e.g., the *student* role of a person, the *taxi* role of a car, the *market* role of an open space), which are optional for their bearer. Disposition describes the tendency in an entity toward something (e.g., a magnet to produce an electrical field), which is not optional for its bearer. State describes the condition of other entities, with respect to certain circumstances, like illness, happiness, sitting, and knowing. There are no states in BFO, and DOLCE classified them under stative perdurant. However, because they do not imply changes they cannot be processes or events.

---

[6] BFO does not support social agents per se, however it was suggested by Smith (2012) to classify organizations as object aggregates under immaterial entity; but some BFO domain ontologies (e.g., see the Ontology of Medically Related Social Entities) suggested to classify organizations under material entity, and noted that they were also thinking to classify it under as a dependent continuant.

*Qualities* (like size, color, mass, or length) inhere in other entities to describe, distinct and express them. Following DOLCE, we differentiate between qualities (e.g. color) and values of qualities (e.g., red). Values of qualities are either *attributes* or *quantities*, which are abstract entities. For example, if the mass of my laptop is 2kg, it means that *mass* is a quality of my laptop that has the value (i.e., quantity) of 2; similarly, "my laptop is heavy", heavy is an attribute of my laptop, i.e., the value of the mass of my laptop is heavy. BFO does not differentiate between qualities and attributes, thus the "mass 2k" is a quality of my laptop.

*5.3. Abstracts {مجرد}*

*Abstract* is an entity that exists only in mind, cannot be measured, and has no location in place or time (e.g., the fairness of *this* judge is *fair*, the color of *this* flower is *red*, number 10, or the set {1,2}). Instances of abstract can be distinguished if they are intentionally realized in mind to be the same/different instances, independently of their embodiment; for example, 10 and 90/9 are the same number. This notion of abstract corresponds to abstract in DOLCE and SUMO. We currently have six subtypes of abstracts: quantity, attribute, formal entity, proposition, and description, which are not elaborated yet except the formal entity branch that consists of about 50 subtypes of sets and numbers.

*5.4. Information Entities {تعبير}*

We adopted this notion from DOLCE+DnS, and revised its definition to be an entity that relationally represents other entities. It bears informational content independently from how this content is recognized. Information realization represents the concrete realization of information (like this copy of the article you are reading), while an information object is the information itself independently from how its content is realized (e.g., the content of the article before it was printed).

**6. Evaluation**

This section evaluates how much the current version of the ontology top-levels is comprehensive and is able to *top* (i.e., be supertype of) the concepts of the Arabic terms, which we call a *comprehensiveness evaluation*. Given a concept of an Arabic term, we would like to know whether it can always be placed as a subtype of (or as equivalent to) a concept in the ontology. Ideally, every concept of an Arabic term should be placed either as equivalent to a node in the ontology, or as a subtype of a *leaf node*. Placing a concept as equal to a node means that this concept is already included, and placing it as under a leaf node means that we have a place for it (its supertype) and can be included later. If an Arabic concept cannot be placed as equal to a node, or under a leaf node in the ontology, it means that the ontology is missing some general categories that should be included in the upper levels. Thus, the lower a concept is classified is an indication that the more comprehensive the ontology will be.

As explained in Section 5, the top-level categories are methodologically important since they are also used to help deriving lower concepts. Therefore, having comprehensive top categories at this early phase will help in extending the ontology in later phases. Quantifying the *comprehensiveness* of the ontology is not easy as it requires collecting and classifying all possible concepts that Arabic can lexicalize, which is not possible because the number of concepts in a language is typically unlimited. Nevertheless, our aim here is rather to have a relative image about how much the top categories in the ontology are general enough to top most Arabic concepts, and to know whether some important top categories need to be included at this phase.

To conduct such an evaluation, we had to find a set of Arabic concepts that are general enough and diverse enough to represent, to a certain extent, most Arabic concepts, and to experiment classifying them under the ontology. For this task, we chose to classify about 2100 general concepts that we found in an old philosophy lexicon called The Definitions "Taʿrīfāt" (Ibn Mohammed, 1983). The lexicon was written by Al-Jurjānī (1339–1414 AD), who is a known philosopher, astronomer, and a linguist. Al-Jurjānī's definitions are written in a glossary style (i.e., a term, with a few lines definition) and his list of terms covers the most abstract notions used by scientists, philosophers, astronomers, and theologians of that time. As will be discussed later, although some of his concepts and terms might be obsolete, the large majority are still communicated and used nowadays (see next Section 6.2). Nevertheless, our evaluation is relative to this lexicon, which we found the best attainable choice to conduct our experiment, and as we could not find a better modern resource covering the most abstract concepts in Arabic.

*6.1. Experiment Set up*

The 2100 definitions of Al-Jurjānī were given to two postgraduate students who are well-trained and familiar with the Arabic Ontology. Each person was asked to separately (1) understand every definition and to classify it (as equivalent to or as subclass of) a node in the ontology, and to (2) specify two measures: *precision* and *confidence,* for each mapped tuple. As explained in Subsection 7.3, precision refers to how much a person believes that this concept is equivalent/subclass of the mapped ontology node. Confidence refers to how much a person is confident about his/her decision. After finishing all mappings, the two results were compared. The set of different mappings, as well as mappings with low precision or confidence were identified. That is, we used the notions of precision and confidence here as a methodology to catch weak and uncertain mappings, which are not many, as will be discussed in the next section. This set of different and uncertain mappings were then jointly discussed by the two persons and a third expert, in long face-to-face meetings, and a consensus is reached about each case, as will be discussed in the next section.

*6.2. Results*

The final results of the experiment are summarized in Figure 6. The experiment included other lower levels of the ontology, but they are not included here for brevity. The number placed at the *top-right* of each node (in blue) is the number of equivalent concepts, and the number *under* each node (in green) is the number of its subclasses. For example, for abstract, there were 2 concepts mapped as equivalents, and 9 as subclasses – that we could not classify under any of the five subclasses of abstract. Similarly, 2 concepts were mapped as equivalent to object, but no subclasses were found; 30 concepts are classified under physical object, and so on.

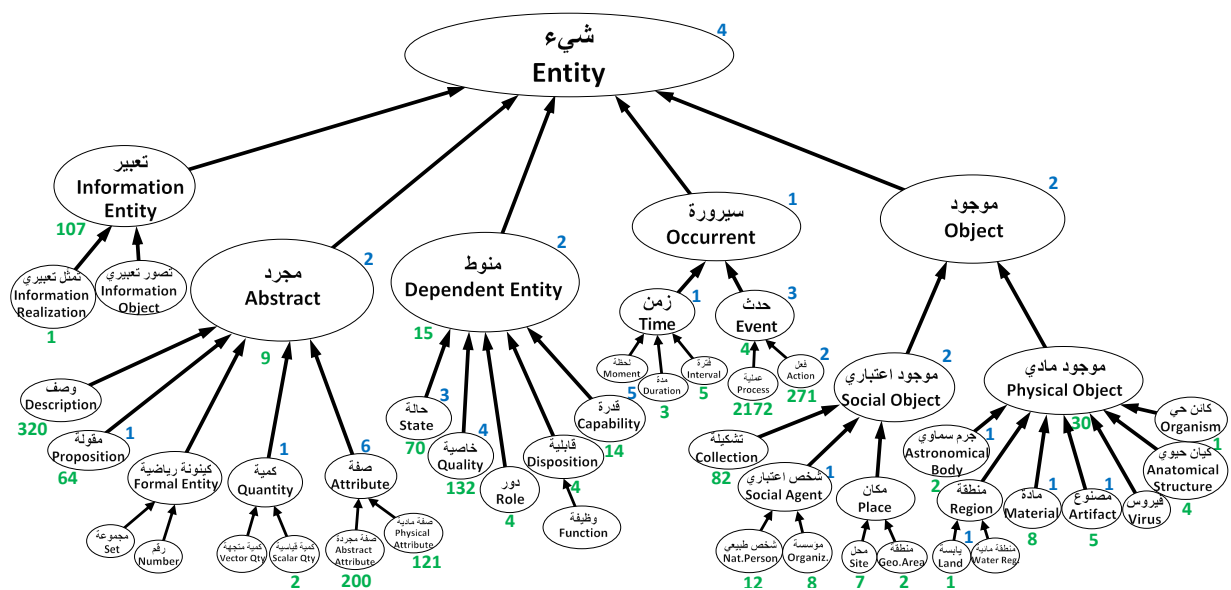

**Figure 6.** Summary of the experiment results

There are about 270 concepts that we excluded from the experiment because we could not map them to the ontology, mainly because (i) we could not understand their meanings, as they are written in an old classical Arabic and without sufficient explanation, or because (ii) they are not concepts in the logical sense, i.e., not classes that can be instantiated, such as الأبد (forever), أصول الفقه (religious doctrine foundations), الأفق الأعلى (the top of a spiritual horizon), الأعيان الثابتة (Immutable Entities – a philosophical religious concept), and others.

Among the remaining 1830 mapped concepts, there are 40 concepts mapped as equal (i.e., the blue numbers) and 1615 concepts mapped as subclasses of leaf nodes. This means that, together, there are 1655 concepts that are correctly placed in the ontology, which is 90% of the mapped concepts. The other 156 concepts (10%) that are mapped as subclasses of non-leaf nodes illustrate cases of missing top categories in the ontology. These are the 30 concepts under physical object, 4 under event, 15 under dependent entity, 9 under abstract, and the majority 107 under information entity.

As explained above, our aim is not to quantify the exact comprehensiveness of the ontology but rather to have a relative idea about missing top categories. If Al-Jurjānī's definitions are to represent Arabic, then most of the missing categories are information entities and linguistic categories like verb, noun, and nickname. In fact, we found many of the missing categories not to be commonly used notions (e.g., glow, flame, lightning, treasure, angel, deposit, and heritage); but there are also some commonly used categories that we currently miss (like light, sound, machine, domain, and others) and should be added to the ontology in the next phase.

In what follows we give a short analysis of the (dis)agreement during the mapping process, which we quantified in Table 1. The persons involved in the mapping are called A1 and A2, while 'Reference' refers to the mappings that were jointly discussed and agreed with a third expert. The first row in the table compares the mappings between A1 and A2, in which they agree exactly on 784 (i.e. 50%), partially agree on 291, and disagree on 481 mappings; and there are 540 cases that were not mapped by A1 or A2, because they were totally not understood by them, as explained above. The second and third rows compare A1 and A2 with the reference mappings. Partial agreement refers to some special-case mappings, which are: 'attribute' vs. its two subtypes, 'property' vs. its three subtypes, 'collection' vs. 'community', and 'event' vs. 'action'. For example, A1 maps a concept to 'attribute' while A2 maps it to 'physical attribute' that is a direct subtype of 'attribute''. Nevertheless, the majority of these partial mappings were related to 'description' vs. 'proposition'. Interestingly, we found that most of the different mappings (i.e., disagreements) were between these and other categories in the ontology. One reason of these partial and different agreements could be that the difference between these ontological categories might not be intuitive, especially that they are too abstract. Another reason, from the lexicon side, and which we feel is most likely, is that the mapped concepts were not easy to understand in the first try, but were better understood when they were deeply discussed with the third expert. If exact and partial mappings are combined, then the agreement would be 69%, 78%, and 82% in the three columns, which generally illustrates high agreements and smoothness in the mapping process. Nevertheless, a full evaluation of the intuitiveness of the ontological categories is not our target and beyond the scope of this article.

|  | Mapped | | | Couldn't Map |
|---|---|---|---|---|
|  | **Exact Mapping** | **Partial Mapping** | **Different Mapping** |  |
| A1 vs. A2 | 784 (50 %) | 291 (19%) | 481 (31%) | 540 (35%) |
| A1 vs. Reference | 1175 (66%) | 215 (12%) | 390 (22%) | 316 (18%) |
| A2 vs. Reference | 1218 (71%) | 187 (11%) | 305 (18%) | 386 (23%) |

**Table 1.** Summary of the (dis/)agreement in the mapping process.

## 7. Ontology Portal and Use Case (mapping and enriching Arabic lexicons)

This section illustrates our lexicographic search engine (Jarrar & Amayreh, 2019, Alhafi et al., 2019), which includes the ontology, as well as about 150 Arabic multilingual lexicons that are being mapped and semantically enriched using the ontology. The ontology and the lexicons can be fully searched and browsed through the search engine. Users may browse the ontology tree as shown on the right-hand side of Figure 7. Each division in the webpage represents a concept: the first line contains the Arabic and English synsets, then the glosses and an example sentence; the last line, in green, contains the *conceptID*, *ConceptProfile* icon (), and the semantic relationships with other concepts (such as *TypeOf*, *Parts*[7], and *Instances*). By clicking on the *ConceptProfile icon* a window appears, which contains the full description of the concept, formal axioms, and its mapping with other resources.

---

[7] Although there are different types of parthood relations (e.g., *temporal-part-of*, *occurrent-part-of, etc.*), they are all called "*PartOf*" in the portal for simplicity and human use.

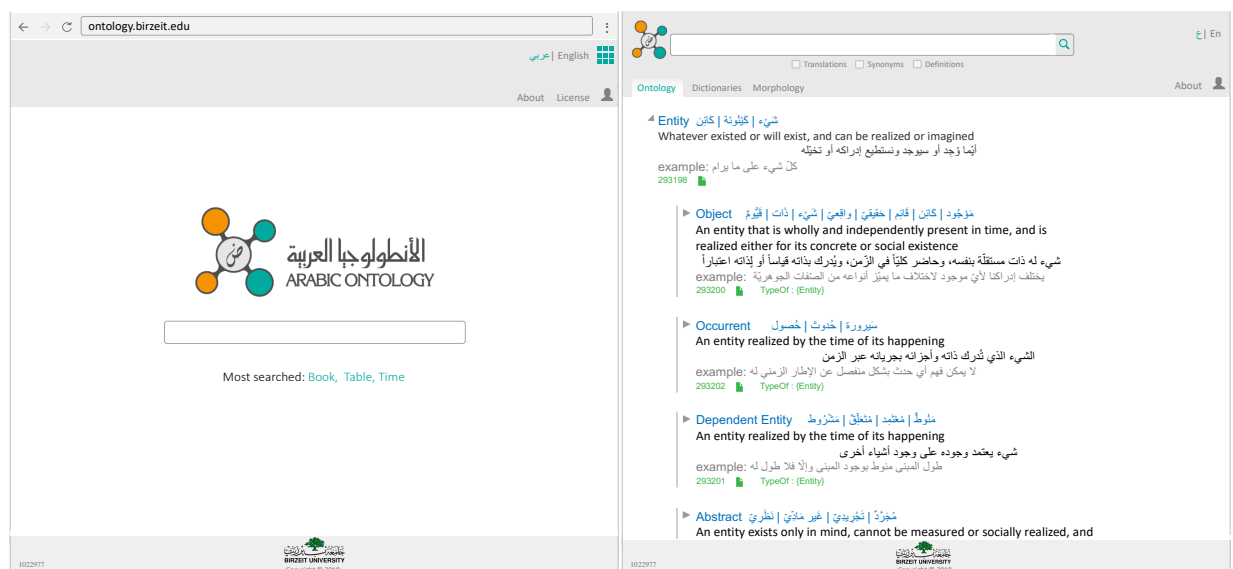

**Figure 7.** Illustration of the ontology and the Lexicographic Search Engine

*7.1. URLs Design*

For interoperability purposes, and to allow the ontology to be linked with other resources in the Linguistic Linked Open Data Cloud (Chiarcos et al., 2011), the URLs in the ontology are designed according to the W3C's Best Practices for Publishing Linked Data, as explained in the following schemes:

- *Concepts*: each concept is given a URL based on its unique ConceptID: `http://{domain}/concept/{ConceptID}`.

- *Semantic relations*: the semantic relations for a given concept can be accessed through URLs: `http://{domain}/concept/{Relation}/{ConceptID}`, to allow one to retrieve, e.g. the instances, of a given conceptID.

- *Terms*: each term is given a URL: `http://{domain}/concept/{term}`, which refers to the set of concepts that are lexicalized using a certain term, i.e., that have this *term* among their synsets.

**Example URLs:**
```
http://ontology.birzeit.edu/concept/293254
http://ontology.birzeit.edu/concept/instances/293121
http://ontology.birzeit.edu/concept/parts/293121
http://ontology.birzeit.edu/concept/virus
```

*7.2. Arabic Multilingual Lexicons*

The search engine includes also about 150 Arabic multilingual lexicons (Jarrar & Amayreh, 2019), for which we obtained a permission to publish in the search engine. We have collected these lexicons, digitized them from scratch, cleaned and structured their content, and stored them in a normalized database schema (Amayreh et al., 2019). Most of these lexicons are multilingual, some are general whereas others are domain-specific. To enrich the lexicons and semantically tag them with rigorous ontological knowledge, we link the lexical senses in each lexicon with a concept in the ontology (Jarrar et al., 2019). In what follows, we introduce the mapping framework, and discuss our progress in the mapping process.

*7.3. Mapping Framework*

This framework aims to enable entities across different linguistic resources to be semantically interlinked through a mapping correspondence, whether these entities are concepts, lexical senses, or individuals.

**Definition:** Given two entities $e_1$ and $e_2$, a *mapping correspondence* between them is defined as:
$$< e_1, e_2, R, P, C>$$

where:
- $e_1$ and $e_2$ are the entities to be mapped, which might be concepts, individuals, or lexical senses.
- $R$ is the mapping relationship between $e_1$ and $e_2$, such as `SameAs, Subtype,` or `InstanceOf`.
- $P$ is the *precision* of the mapping.
- $C$ is the *confidence* degree.

The precision and confidence degrees are introduced for methodological purposes – to indicate weak or doubtful mappings (see a similar framework by Bouquet et al., 2004). For example, `<a,b,SameAs,95%,70%>` means that `a` and `b` are 95% the same, and that we are 70% sure that this mapping is correct. In other words, precision refers to how much the mapper (human or software) believes that the first entity is related to (e.g., SameAs) the second entity. Although this might be interpreted in a set-theoretic semantics to mean e.g., that 95% of the extensions of two concepts are the same, we introduced this informal notion to document only some weak mappings with lexical senses that might not be vague anyway. The notion of confidence refers to how much the mapper (human or software) is confident that the mapping is correct or reliable. As discussed in the previous section, these notions are methodologically helpful to identify weak and doubtful mappings in order to exclude or correct them.

This framework is used to map concepts in the Arabic Ontology and lexical senses and entities in other resources. The ontology is fully mapped with WordNet, Wikidata, and with about 13445 lexical senses from different Arabic lexicons. The mapping process was done manually using the above mapping framework. In these mapping we only used simple mapping relations, as shown in Table 2. Complex mapping correspondences and rules are not considered at this stage. Table 2 provides some statistics about our progress in the mapping process, and which mapping relations are used. In this way, we plan to continue mapping all lexicons to the ontology, as well as to each other, in order to build an Arabic big linguistic data graph enriched with lexical and semantic features. Jarrar et al. (2019, 2018) discussed and illustrated further how these mappings are being used to represent Arabic lexicons in the W3C Lemon Model and link them with external sources, which is an emergent linked data paradigm (Chiarcos et al., 2011; Gracia et al., 2014). The mapping and Lemon representation of each lexical concept can be accessed by clicking on the RDF icon displayed beside each of the retrieved results.

| Relation | Number of Mappings |
|---|---:|
| `SameAs` | 11400 |
| `SubClassOf/SuperClassOf` | 1050 |
| `PartOf/HasPart` | 100 |
| `InstanceOf/Type` | 770 |
| `Similar` | 125 |
| **Total** | 13445 |

**Table 2.** Statistics about our progress in the mapping process.

## 8. Conclusions and Future Work

We have proposed an ontology for Arabic and explained its foundations and the methodology we followed to develop it, stressing the issue of benchmarking concept definitions and classification to scientific advances and what scientists and subject matter experts agree upon, rather than what native speakers naïvely accept. The notion of concept was also discussed and clarified, and a strict methodology for gloss formulation was proposed. The content and the top-level concepts of the ontology were presented, and their comprehensiveness was evaluated. The ontology web portal is illustrated and a use case of mapping it to many Arabic lexicons is presented.

Despite the fact that the ontology currently contains about 1,300 completed concepts and another 11,000 concepts that are partially ready, there are still many open issues that need to be resolved. There are several branches that need to be extended and elaborated, such as information objects, attributes, dependent entities, qualities, events and processes, and others. Events and processes need special care, as they are hard to ontologically analyze. Furthermore, since there are many scientific ontologies developed in English, especially in natural sciences, as well as many scientific and specialized glossaries available in Arabic, we believe that such ontologies, glossaries and classifications, can be revised and reused. Providing a full mapping into SUMO and other resources will increase the usability of our ontology. Among other issues we plan to address, is linking the ontology with known Arabic corpora (Maamouri et al, 2020; Jarrar et al, 2014, Jarrar et al, 2016), which provides great utility for several NLP applications. This may not be very complicated, as we used the LDC's SAMA lemmas that are linked with the Arabic corpora.

### Acknowledgments

We would like to acknowledge the input provided by Rana Rashmawi during her employment and master studies. We would like to also thank many colleagues and students who contributed to the ontology during the discussion sessions over the past years, which helped us improve the ontology in many ways, specially Prof. Wassim Abu Fasha, Prof. Jamal Daher, Hiba Olwan, Alaa Bazzar, and others, as well as Werner Ceusters for his suggestions and proofreading early drafts of the article.